\documentclass[letterpaper, 10 pt, conference]{ieeeconf}  

\IEEEoverridecommandlockouts                              

\usepackage{graphics} 
\usepackage{epsfig} 
\usepackage{mathptmx} 
\usepackage{times} 
\usepackage{amsmath} 
\usepackage{amssymb}  
\usepackage{algorithm}
\usepackage{algpseudocode} 
\usepackage{booktabs}
\usepackage[table]{xcolor}

\title{\LARGE \bf
Task-Level Decisions to Gait Level Control: A Hierarchical Policy Approach for Quadruped Navigation
}

\author{Sijia Li, Haoyu Wang, Shenghai Yuan, Yizhuo Yang, Thien-Minh Nguyen$^{*}$
\thanks{$^{*}$ Corresponding author, email: thienminh.nguyen@uq.edu.au}%
}

\begin{document}

\maketitle
\thispagestyle{empty}
\pagestyle{empty}

\begin{abstract}
Real-world quadruped navigation is constrained by a scale mismatch between high-level navigation decisions and low-level gait execution, and by instabilities under out-of-distribution environmental changes. Such variations challenge sim to real transfer and can trigger falls when policies lack explicit interfaces for adaptation. In this paper, we present a hierarchical policy architecture for quadrupedal navigation. A low-level policy, trained with reinforcement learning in simulation, delivers gait-conditioned locomotion and maps task requirements to a compact set of controllable behaviour parameters, enabling robust mode generation and smooth switching. A high-level policy makes task-centric decisions from sparse semantic or geometric terrain cues and translates commands into low-level targets, forming a traceable decision pipeline without dense maps or high resolution terrain reconstruction. Different from the end-to-end approach, our architecture provides explicit interfaces for deployment time tuning, fault diagnosis, and policy refinement. We introduce a structured curriculum with performance-driven progression that expands environmental difficulty and disturbance ranges. Experiments show higher task success rates on mixed terrains and out of distribution tests.
\end{abstract}

\section{INTRODUCTION}

Real-world quadruped navigation is a key capability for bringing legged robots from controlled settings to open-world deployment, with significant value for outdoor inspection and maintenance, emergency response, mobile operations in industrial sites, and autonomous field exploration \cite{hoeller2024anymal}. The challenge in real environments is not only to maintain dynamic stability, but also to continuously make appropriate action choices under task constraints and reliably reach the goal. End-to-end task success requires consistent coordination between task intent and locomotion execution within the same closed loop. This requirement becomes particularly critical when environmental information is sparse and incomplete and contact dynamics are highly uncertain \cite{focchi2019heuristic}. Focusing solely on low level stabilizing control often fails to ensure efficient task progression, while relying only on high level planning cannot promptly absorb the dynamic effects of local contacts and transient disturbances. Therefore, a central challenge in quadruped navigation remains how to integrate information across abstraction levels within a unified control loop and to form a deployable and tunable decision and control pipeline \cite{fu2022coupling}.

Existing work on navigation and control system design has developed several representative paradigms. Classical perception localization mapping and planning pipelines generate feasible trajectories from relatively complete environment representations and track them using model based control or optimization at the execution layer \cite{grandia2023perceptive,fahmi2022vital}. These pipelines provide a clear engineering structure, yet real deployment often requires high quality sensing and high resolution environment modeling, leading to complex system integration. Errors can propagate across modules and trigger cascading failures, and closed loop responsiveness to rapidly changing conditions is limited. End to end learning methods, especially imitation learning based mappings from observations to actions, reduce the burden of explicit modeling and modular design, but typically depend on large scale expert demonstrations with broad coverage \cite{cheng2024extreme}. Data collection is expensive and cannot cover long tail operating conditions, which limits stability and generalization under out of distribution settings and leaves few structured interfaces for deployment time tuning diagnosis and correction \cite{zhang2024resilient}. In parallel, robust learning based low level gait controllers and local reactive strategies can improve disturbance rejection and local traversability \cite{lee2020learning}, but without task consistent high level constraints and executable cross layer coordination, they often struggle to support long horizon task execution, leading to behavior drift policy conflicts or low task efficiency \cite{margolis2024rapid}. Overall, achieving a balanced solution across deployment feasibility out of distribution robustness and task level closed loop consistency still calls for a hierarchical policy framework that is easy to integrate debug and maintain \cite{jain2019hierarchical}.

To address these challenges, we propose a deployable hierarchical policy architecture in which task level decision making and gait level execution operate collaboratively within the same control loop and are linked through explicit interfaces that enforce a consistent mapping from decisions to executable commands. The low level policy is trained with reinforcement learning in simulation to learn gait conditioned locomotion control, and modulates locomotion modes using a compact and controllable set of behavior parameters, enabling robust generation and smooth switching across multiple modes. The high level policy is task centric and, at each decision step, generates commands from available semantic or geometric terrain cues, then directly converts them into low level executable behavior parameters and control targets. This design yields a structured and debug friendly decision and control pipeline without relying on dense maps or high resolution terrain reconstruction. For training, we introduce a structured curriculum learning mechanism that progressively expands environmental difficulty and disturbance ranges through performance driven advancement to improve training efficiency and cross terrain robustness. Experimental evaluations show that the proposed framework achieves higher task success rates on mixed terrains and out of distribution tests. Our main contributions are threefold.
\begin{enumerate}
    \item We present a synchronized hierarchical policy system that couples task-level decisions and gait-level execution within a unified closed loop through explicit cross-layer interfaces, mitigating performance degradation caused by scale mismatch at the system level.
    \item We propose gait-conditioned low-level control with compact behavior parameterization, which enables a stable mapping from task commands to executable low-level targets, supports robust generation and smooth switching across locomotion modes, and provides direct mechanisms for rapid deployment-time tuning, fault diagnosis, and policy correction.
    \item We introduce a performance-driven structured curriculum training pipeline that improves training efficiency and cross-terrain generalization, leading to higher task success rates on mixed terrains and out-of-distribution evaluations.
\end{enumerate}

\section{RELATED WORKS}

\subsection{Learning Based Locomotion Control for Quadrupeds}
Learning-based methods have significantly advanced locomotion control for quadrupedal robots, enabling robust and adaptive behaviors that are difficult to achieve with classical model-based approaches \cite{zhang2022deep, kotha2024next}. Reinforcement learning (RL) has emerged as a dominant paradigm, allowing policies to be trained in simulation and transferred to real hardware through careful domain randomization and reward shaping \cite{miki2022learning, choi2023learning}. Recent work has demonstrated that RL-trained controllers can handle a wide range of terrains and external disturbances by leveraging rich proprioceptive feedback \cite{xiao2024paloco, kim2024not}. Further improvements in gait quality and energy efficiency have been achieved through structured action representations, phase-based motion priors, and multi-gait parameterization \cite{bellegarda2022cpgrl, chen2023learning}. Generalization across morphologically diverse platforms and task conditions has also been explored, producing controllers that transfer broadly without per-robot retraining \cite{shafiee2024manyquadrupeds}. More recently, combining RL with generative pre-trained models has enabled lifelike and agile behaviors, while adaptive controllers have been proposed to handle challenging and deformable contact conditions \cite{luo2024moral}. These developments have established a strong foundation of transferable low-level execution primitives, which our work builds upon by coupling such controllers with a task-level decision layer.

\subsection{Task-Level Navigation for Legged Robots}
Navigation and high-level decision making for legged robots have been approached from both classical planning and learning-based perspectives. Classical pipelines typically construct explicit environment representations and generate feasible trajectories tracked by model-based controllers \cite{wellhausen2023artplanner, shamsah2023integrated}. While these pipelines offer interpretable structure, they require high-quality sensing and dense reconstruction, and error propagation across modules can limit reliability in unstructured environments \cite{deluca2023autonomous}. Abstraction-based and uncertainty-aware planning methods have been proposed to improve robustness under partial observability \cite{jiang2023abstraction}. Legged systems have also been deployed in demanding field scenarios such as planetary analog exploration \cite{arm2023scientific}, highlighting the practical need for reliable long-horizon navigation. For wheeled-legged platforms, end-to-end learning of navigation and locomotion has shown promising results in unstructured outdoor settings \cite{lee2024learning}. Our work draws on these insights and adopts sparse terrain representations at the task level to support robust long-horizon navigation without relying on dense maps or full environment reconstruction.

\subsection{Hierarchical Policies and Training Strategies}
Hierarchical policy architectures have been widely studied as a means to decompose complex control problems into sub-tasks operating at different time scales \cite{zhu2022hierarchical, jin2022learning}. In robotic manipulation, hierarchical designs have demonstrated strong performance on contact-rich \cite{simonic2024hierarchical}, multi-object search \cite{schmalstieg2023learning}, and multi-task settings \cite{ma2024hierarchical}, as well as in complex sequential task solving \cite{triantafyllidis2023hybrid} and high-precision domains such as autonomous surgery \cite{kim2025srth}. In the context of legged locomotion, hierarchical designs separate high-level goal-directed behavior from low-level motor control, communicating through explicit or implicit command interfaces \cite{han2024lifelike}. Curriculum learning has further been shown to improve training robustness by progressively exposing agents to increasing task difficulty, reducing the risk of policy collapse and improving generalization \cite{feng2023genloco}. Our framework integrates these ideas into a unified hierarchical architecture with an explicit cross-layer command interface and a structured performance-driven curriculum, targeting improved robustness and deployability on mixed and out-of-distribution terrains.

\begin{figure*}[t]
  \centering
  \includegraphics[width=\textwidth]{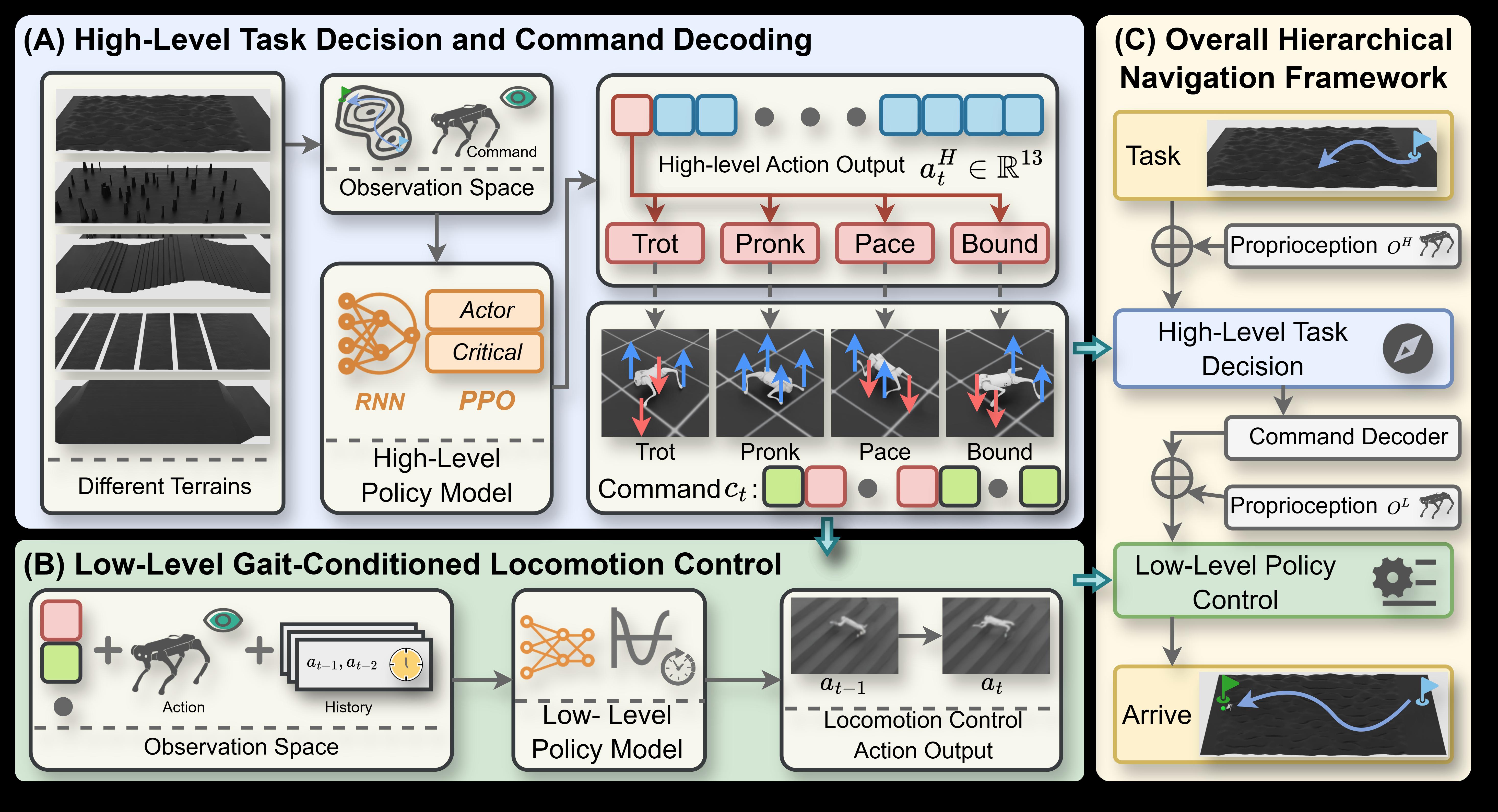}
  \vspace{-15pt}
    \caption{Hierarchical navigation framework. (\textbf{A}) The high-level recurrent policy maps task/terrain cues to a compact behavior output, which is decoded into an executable command and a discrete gait selection (trot, pronk, pace, bound). (\textbf{B}) The low-level gait-conditioned policy uses proprioception, the decoded command, and short action history to generate joint-level actions for stable locomotion. (\textbf{C}) The two policies run in closed loop with proprioceptive feedback to support goal-reaching across diverse terrains.}
  \label{fig:hierarchical_framework}
    \vspace{-15pt}
\end{figure*}

\section{PRELIMINARIES}
\subsection{Task Setup and Terrain Curriculum}
We study episodic quadruped navigation on a curriculum of procedurally generated terrains. The environment is organized as a two dimensional grid whose rows correspond to difficulty levels and whose columns correspond to terrain types. In each episode, we sample one terrain tile from this grid, place the robot at a designated start region, and specify a goal region within the same tile. The episode objective is to reach the goal and maintain a stable posture. To ensure a consistent notion of difficulty across terrain types, we introduce a normalized difficulty variable $d \in [0,1]$ and use a shared affine mapping to control terrain specific geometric parameters
\begin{equation}
p(d) = p_{\min} + d \left(p_{\max} - p_{\min}\right).
\end{equation}
All terrains are represented as height fields $h(x,y)$. We include flat safety regions near the start and the goal, and we add a transition band between safety and challenge regions to avoid abrupt height changes that can destabilize learning and evaluation.

The curriculum includes five terrain families, named \textit{Rough}, \textit{Pillar}, \textit{Stair}, \textit{Gap}, and \textit{Tilt}. \textit{Rough} terrains introduce continuous surface irregularities using a noise based height field, where the disturbance magnitude is governed by an amplitude schedule $A(d)$
\begin{equation}
h(x,y) = A(d)\,\tilde{n}(x,y),
\end{equation}
and $\tilde{n}(x,y)$ denotes a zero mean normalized noise field. \textit{Pillar} terrains place a set of cylindrical obstacles on a flat base, and the primary difficulty driver is the number of obstacles $N_{\mathrm{obs}}(d)$, which increases with $d$. \textit{Stair} terrains create periodic elevation changes, and difficulty is controlled by the step height $h_s(d)$, which increases with $d$. \textit{Gap} terrains remove support regions to form cracks and holes. The resulting support domain can be expressed as
\begin{equation}
\Omega_{\mathrm{support}}(d) = \Omega \setminus \bigcup_i \mathcal{G}_i(d),
\end{equation}
where $\Omega$ is the planar terrain domain and $\mathcal{G}_i(d)$ denotes the removed region of the $i$th gap whose width increases with $d$. \textit{Tilt} terrains apply a lateral slope to induce a persistent sideways component of gravity. A representative height profile is
\begin{equation}
h(y) = \left(y - \frac{W}{2}\right)\tan\theta(d),
\end{equation}
where $W$ is the terrain width and the tilt angle $\theta(d)$ increases with $d$. This curriculum provides a controlled expansion of contact uncertainty and disturbance magnitude under a shared difficulty variable, enabling consistent training and evaluation on mixed terrains and out of distribution conditions.

\subsection{MDP Formulation}
We formalize the hierarchical navigation controller as two Markov decision processes that operate at different time scales and interact through an explicit command interface. The execution level process is defined as $\mathcal{M}_L=(\mathcal{S}_L,\mathcal{A}_L,\mathcal{P}_L,\mathcal{R}_L,\rho_0^L,\gamma)$. The state space $\mathcal{S}_L$ is dominated by proprioceptive information and is conditioned on commands provided by the high level policy. The action space $\mathcal{A}_L$ is continuous and produces joint level control outputs. The transition kernel $\mathcal{P}_L$ is induced by the robot dynamics and contact interactions, and the reward function $\mathcal{R}_L$ shapes locomotion stability and tracking quality. The initial state distribution is $\rho_0^L$ and the discount factor is $\gamma$.

The decision level process is defined as $\mathcal{M}_H=(\mathcal{S}_H,\mathcal{A}_H,\mathcal{P}_H,\mathcal{R}_H,\rho_0^H,\gamma)$. The state space $\mathcal{S}_H$ contains task relevant variables together with standard proprioceptive signals. The action space $\mathcal{A}_H$ is a compact set of behavior parameters that are decoded into executable commands for the execution layer. The reward function $\mathcal{R}_H$ is designed to reflect task progress and stable goal reaching. Since each decision step triggers closed loop execution over multiple physics steps, the induced transition kernel $\mathcal{P}_H$ is determined jointly by the command decoder, the frozen execution policy, and the underlying dynamics integration.

\section{METHODOLOGY}
This section presents the proposed hierarchical policy and its training procedure. As shown in Fig.~\ref{fig:overview}, we couple task-level decision making and gait-level execution through an explicit command interface, which constrains long-horizon choices to a dynamically feasible command space. The system consists of a high-level task policy $\pi_H$, a command decoder $\mathcal{D}$, and a low-level gait-conditioned controller $\pi_L$. At each control step, the high-level policy maps task observations to a compact behavior parameter vector, the decoder converts it into an executable command, and the low-level controller produces joint-level actions conditioned on this command and proprioceptive observations. The resulting closed-loop policy can be written as
\begin{equation}
a_t^{H}=\pi_H(o_t^{H}),\qquad
c_t=\mathcal{D}(a_t^{H}),\qquad
a_t^{L}=\pi_L(o_t^{L},c_t).
\end{equation}
Both policies are optimized with reinforcement learning, while their roles are complementary. The low-level policy learns stable gait generation and command tracking, and the high-level policy learns task progression under this executable interface. The remainder of this section first describes the low-level controller, then the high-level policy with reward design, and finally a unified algorithm for training and execution.

\subsection{Low-Level Gait-Conditioned Controller}
The low-level controller $\pi_L$ converts the command $c_t$ into stable joint-level control actions, providing a consistent execution interface for the high-level policy under contact uncertainty and external disturbances. To support multi-modal locomotion and smooth transitions, the controller is conditioned on a discrete gait index $g_t\in\{0,1,2,3\}$ and a phase clock vector $\phi_t$. The four gaits used in this work are trot, pronk, pace, and bound. The low-level observation is dominated by proprioceptive signals and is augmented with the command and gait conditions, which we write as
\begin{equation}
s_t^{L}=\big[g_t^{b},\ \Delta q_t,\ \dot q_t,\ a_{t-1},\ a_{t-2},\ c_t,\ \phi_t,\ g_t\big],
\end{equation}
where $g_t^{b}$ encodes base orientation relative to gravity, $\Delta q_t$ and $\dot q_t$ denote joint position error and joint velocity, and $a_{t-1},a_{t-2}$ are action history terms. The controller outputs a continuous action vector $a_t^{L}\in\mathbb{R}^{12}$, which is mapped to joint position targets through a clipped and scaled interface
\begin{equation}
\bar a_t^{L}=\mathrm{clip}(a_t^{L},-a_{\max},a_{\max}),\qquad
q_t^{\star}=q_0+\alpha\,\bar a_t^{L}.
\end{equation}
This mapping bounds the actuation magnitude and yields a predictable execution channel, which reduces cross-layer coupling during training.

The low-level reward shapes gait quality and command tracking while discouraging excessive energy use and high-frequency jitter. The reward construction follows the design in \cite{margolis2023walk} and uses a shared set of core terms across the four gaits
\begin{equation}
R_t^{L}=\sum_{i} w_i\, r_i(s_t^{L},a_t^{L}).
\end{equation}
The key terms include tracking of planar velocity and yaw rate, body stabilization, action smoothness, and energy regularization. As an example, tracking terms can be expressed using exponential errors
\begin{equation}
r_t^{\mathrm{lin}}
=
\exp\!\left(
-\frac{\left\lVert v_{xy}-v_{xy}^{\mathrm{cmd}}\right\rVert_2^{2}}
{\sigma_{\mathrm{lin}}}
\right),
\end{equation}
\begin{equation}
r_t^{\mathrm{yaw}}
=
\exp\!\left(
-\frac{\left(\omega_z-\omega_z^{\mathrm{cmd}}\right)^{2}}
{\sigma_{\mathrm{yaw}}}
\right).
\end{equation}
The low-level policy is learned by maximizing the expected discounted return
\begin{equation}
\pi_L^{\star}=\arg\max_{\pi_L}\ \mathbb{E}\!\left[\sum_{t=0}^{\infty}\gamma^t R_t^{L}\right].
\end{equation}
This yields a transferable set of gait-conditioned execution primitives that robustly track high-level commands and provide the foundation for high-level task learning.

\subsection{High-Level Task Policy}
The high-level policy $\pi_H$ addresses long-horizon decision making and intent generation. It does not issue joint-level commands. Instead, it outputs a compact behavior parameter vector that is mapped by an explicit decoder $\mathcal{D}$ into an executable command $c_t$ for the low-level controller. This design confines high-level exploration to a dynamically feasible command space, allowing the high-level policy to focus on navigation and mode selection while the low-level policy handles contact stabilization and gait execution. The explicit interface also improves interpretability and supports deployment-time diagnosis and tuning.

\paragraph{Observation and action} At decision step $t$, the high-level policy receives an observation $o_t^H$ that summarizes task progress and robot state. The observation is constructed from task-relevant variables and standard proprioceptive signals, and it avoids reliance on dense maps or high-resolution reconstruction. The policy outputs a $13$-dimensional action $a_t^H\in\mathbb{R}^{13}$ that encodes compact behavior parameters. Since the action is subsequently processed by a deterministic decoder, we first apply elementwise clipping and normalization
\begin{equation}
\tilde a_t^H=\mathrm{clip}(a_t^H,-a_{\max},a_{\max}),
\end{equation}
\begin{equation}
x_t=\frac{\tilde a_t^H}{a_{\max}},
\qquad
x_t\in[-1,1]^{13},
\end{equation}
where $a_{\max}>0$ bounds the action magnitude. This normalization keeps the decoder input within a fixed range and prevents out-of-bound commands that the low-level policy cannot reliably track.

\paragraph{Command decoding and gait discretization} The decoder $\mathcal{D}$ maps the normalized action $x_t$ to a $15$-dimensional command vector $c_t\in\mathbb{R}^{15}$. For continuous command components, we apply an affine range mapping
\begin{equation}
c_{t,j}
=
\ell_j+\frac{x_{t,j}+1}{2}\left(u_j-\ell_j\right),
\qquad
j\in\mathcal{J}_{\mathrm{cont}},
\end{equation}
where $\ell_j$ and $u_j$ are the lower and upper bounds of the $j$th command component and $\mathcal{J}_{\mathrm{cont}}$ denotes the index set of continuous components. This mapping makes the command bounds explicit and supports straightforward deployment-time adjustment.

To enable discrete gait selection and avoid ambiguous continuous switching, a dedicated gait channel is quantized into a four-class gait index $g_t\in\{0,1,2,3\}$
\begin{equation}
g_t
=
\mathrm{clip}\!\left(
\left\lfloor 2\left(x_t^{(g)}+1\right)\right\rfloor,
0,3
\right),
\end{equation}
where $x_t^{(g)}$ is the gait channel and $\lfloor\cdot\rfloor$ denotes the floor operator. The final command concatenates the continuous command components with a gait embedding $e(g_t)$ that encodes the gait-dependent phase template used by the low-level controller
\begin{equation}
c_t=\big[c_t^{\mathrm{cont}},\ e(g_t)\big].
\end{equation}

\paragraph{Reward design} The high-level reward is designed to promote task completion while enforcing stability, safety, and command executability. We use a weighted sum of reward components
\begin{equation}
R_t^{H}
=
\sum_{i\in\mathcal{I}}
\tilde w_i\, r_i(s_t,a_t^{H}),
\end{equation}
where $r_i(\cdot)$ denotes the $i$th reward term, $\tilde w_i$ is its effective weight, and $\mathcal{I}$ is the set of enabled terms. To keep the reward scale consistent across different control step sizes, we apply step-time scaling to all active weights
\begin{equation}
\tilde w_i=w_i\,\Delta t_{\mathrm{step}},
\end{equation}
where $w_i$ is the configured weight and $\Delta t_{\mathrm{step}}$ is the high-level step duration.

The reward contains goal-reaching terms that provide dense gradients and improve convergence near the target. Let $p_t\in\mathbb{R}^3$ denote the base position and let $p_g\in\mathbb{R}^3$ denote the goal position. The distance to the goal is $d_t=\lVert p_t-p_g\rVert_2$. We define a normalized progress variable
\begin{equation}
p=\mathrm{clip}\!\left(1-\frac{d_t}{R_{\mathrm{map}}},0,1\right),
\end{equation}
and use a smooth shaping function
\begin{equation}
r_{\mathrm{goal\_dist}}=p+a\left(1-e^{-bp}\right),
\end{equation}
where $R_{\mathrm{map}}$ is a distance normalization constant and $a>0$, $b>0$ control the curvature. We also encourage facing the goal to reduce lateral drift. Let $\psi_t$ be the current yaw and let $\psi_t^{\star}$ be the desired yaw pointing to the goal. We use
\begin{equation}
r_{\mathrm{face}}=\cos(\psi_t^{\star}-\psi_t).
\end{equation}
To incorporate time efficiency and provide a clear completion signal, we include a fast-arrival bonus inside a goal region of radius $d_0$
\begin{equation}
\alpha_t=\mathrm{clip}\!\left(1-\frac{t}{T_{\max}},0,1\right),
\end{equation}
\begin{equation}
r_{\mathrm{arrive}}=\mathbf{1}[d_t<d_0]\,(b_0+b_1\alpha_t),
\end{equation}
where $t$ is the decision step index, $T_{\max}$ is the episode horizon, and $\mathbf{1}[\cdot]$ is the indicator function. To reduce oscillations after reaching the goal and to match deployment requirements, we reward stable standing within the goal region
\begin{equation}
r_{\mathrm{stable}}
=
\mathbf{1}[d_t<d_0]\,
\exp\!\left(-\eta_t\right)\,
\exp\!\left(-\frac{(z-z^{\star})^2}{\sigma_z}\right),
\end{equation}
\begin{equation}
\eta_t
=
\lVert v^{\mathrm{lin}}\rVert
+
\lVert v^{\mathrm{ang}}\rVert
+
\beta\lVert \dot q\rVert .
\end{equation}
where $v^{\mathrm{lin}}$ and $v^{\mathrm{ang}}$ are base linear and angular velocities, $\dot q$ denotes joint velocities, $z$ is the base height, $z^{\star}$ is a desired height, and $\beta>0$, $\sigma_z>0$ are scaling constants.

We include smoothness regularization because high-level commands drive closed-loop execution. Rapid fluctuations can induce unnecessary mode switching and destabilize contacts. We penalize the high-level action rate
\begin{equation}
r_{\mathrm{action\_rate}}=\sum_j (a_{t}^{(j)}-a_{t-1}^{(j)})^2,
\end{equation}
and penalize the first and second differences of the command vector $u_t$
\begin{equation}
r_{\mathrm{cmd\_sm1}}
=
\frac{1}{n}\sum_{j=1}^{n}(u_t^{(j)}-u_{t-1}^{(j)})^2,
\end{equation}
\begin{equation}
r_{\mathrm{cmd\_sm2}}
=
\frac{1}{n}\sum_{j=1}^{n}\big(\Delta u_t^{(j)}-\Delta u_{t-1}^{(j)}\big)^2,
\qquad
\Delta u_t=u_t-u_{t-1},
\end{equation}
where $n$ is the command dimension.

Safety and degeneracy prevention terms discourage undesired contacts and stagnation. Let $F_j$ be the contact force at contact $j$ and let $F_{\mathrm{th}}$ be a threshold. We penalize excessive contact forces
\begin{equation}
r_{\mathrm{col}}=\sum_j \max(0,\lVert F_j\rVert-F_{\mathrm{th}}).
\end{equation}
To prevent overly conservative solutions that remain stationary far from the goal, we add a laziness penalty
\begin{equation}
r_{\mathrm{lazy}}=\mathbf{1}[d_t>d_0]\ \mathbf{1}[\lVert v\rVert<v_{\mathrm{th}}],
\end{equation}
where $v_{\mathrm{th}}$ is a speed threshold. Finally, we include a constant alive term
\begin{equation}
r_{\mathrm{alive}}=1.
\end{equation}

The high-level policy is trained to maximize the expected discounted return
\begin{equation}
\pi_H^{\star}
=
\arg\max_{\pi_H}\ \mathbb{E}\!\left[\sum_{t=0}^{\infty}\gamma^t R_t^{H}\right],
\end{equation}
where $\gamma\in(0,1)$ is the discount factor. The induced high-level transitions are determined by the decoder, the low-level closed-loop tracking, and the underlying dynamics integration, which ensures that high-level learning remains grounded in executable behavior.

\begin{algorithm}[t]
\caption{High-level training with environment-level curriculum}
\label{alg:overall}
\begin{algorithmic}[1]
\Require Decoder $\mathcal{D}$, window size $W$, threshold $S$, max level $L_{\max}$
\Require Policies $\pi_L(\cdot;\theta_L)$, $\pi_H(\cdot;\theta_H)$, envs $\{E_i\}_{i=1}^{M}$
\State Train $\pi_L$ and export; freeze $\theta_L$
\State Init levels $\{\ell_i\}$ and success windows $\{B_i\}$ of length $W$
\While{not converged}
  \ForAll{$i\in\{1,\dots,M\}$}
    \State $E_i\gets\mathrm{Reset}(\ell_i)$
  \EndFor
  \For{$t=0$ to $T_{\max}-1$}
    \ForAll{$i\in\{1,\dots,M\}$}
      \State $a_{t,i}^{H}\gets\pi_H(o_{t,i}^{H};\theta_H)$
      \State $c_{t,i}\gets\mathcal{D}(a_{t,i}^{H})$
      \State $a_{t,i}^{L}\gets\pi_L(o_{t,i}^{L},c_{t,i};\theta_L)$
      \State $E_i.\mathrm{StepPhysics}(a_{t,i}^{L})$
      \State $(\mathrm{fail}_i,\mathrm{reach}_i)\gets E_i.\mathrm{CheckTermination}()$
      \State $R_{t,i}^{H}\gets E_i.\mathrm{GetReward}()$
      \State $\mathrm{reset}_i\gets \mathrm{fail}_i \lor \mathrm{timeout}_i$
      \If{$\mathrm{reset}_i$}
        \State $\mathrm{success}_i\gets \mathbf{1}[\mathrm{reach}_i]$
        \State $B_i\gets\mathrm{Push}(B_i,\mathrm{success}_i)$
        \If{$B_i\ \mathrm{full}$}
          \State $c_i\gets \sum B_i$
          \State $\ell_i\gets\mathrm{UpdateLevel}(\ell_i,c_i,W,S,L_{\max})$
          \State $B_i\gets\mathrm{Clear}()$
        \EndIf
        \State $E_i\gets\mathrm{Reset}(\ell_i)$
      \EndIf
    \EndFor
    \State $\theta_H\gets\mathrm{PPOUpdate}(\theta_H)$
  \EndFor
\EndWhile
\end{algorithmic}
\end{algorithm}

\subsection{Curriculum Mechanism}
We train the high-level task policy with curriculum learning under a fixed low-level executor. The overall loop connects the high-level policy, the command decoder, the low-level controller, and an environment-level curriculum scheduler. The purpose is to improve task-level robustness by exposing the policy to increasingly difficult terrains while ensuring that exploration remains within a command space that is executable by the low-level controller. Algorithm~\ref{alg:overall} summarizes the full training loop and the per-environment curriculum update.

We adopt a two-stage procedure. First, we train the low-level gait-conditioned policy until it robustly tracks commands across the supported gaits, then export and freeze it for high-level training. Second, we train the high-level policy with the frozen low-level policy injected into the environment, so that policy updates reflect task-level decisions rather than joint-level stabilization. In our implementation, curriculum adaptation is performed independently for each parallel environment.

Let $\ell_i\in\{0,\dots,L_{\max}-1\}$ denote the current terrain difficulty level of environment $i$. We define episode success by whether the agent reaches the goal region at least once within the episode. A reach event is detected when the distance to the goal satisfies $d_{\mathrm{goal}}<0.5$. Episode termination is triggered by failure or time-out. Failure includes falling, detected by base height below $0.15$, and undesired collision, detected by a termination contact force larger than $40$. Time-out is handled separately in the step loop.

Each environment maintains a sliding window of length $W$ over recent episode outcomes. Let $c_i$ be the number of successful episodes in the window. When the window is full, we apply promotion and demotion rules with threshold $S$
\begin{equation}
c_i \ge S \ \Rightarrow\ \ell_i \leftarrow \min(\ell_i+1,L_{\max}-1),
\end{equation}
\begin{equation}
c_i < (W-S+1) \ \Rightarrow\ \ell_i \leftarrow \max(\ell_i-1,0).
\end{equation}
After a level change, the corresponding window buffer is reset for that environment. This environment-level update yields a mixture of difficulty levels within each batch, which improves sample efficiency and reduces overfitting to a single difficulty regime.

Curriculum update is coupled with the environment step in a fixed order. After physics integration and low-level execution, the environment checks termination and reach events, then computes rewards, and finally resets environments whose reset flags are active. This ordering ensures that termination signals and curriculum states are consistent with the reward computed at the same step.

\begin{figure*}[t]
  \centering
  \includegraphics[width=\textwidth]{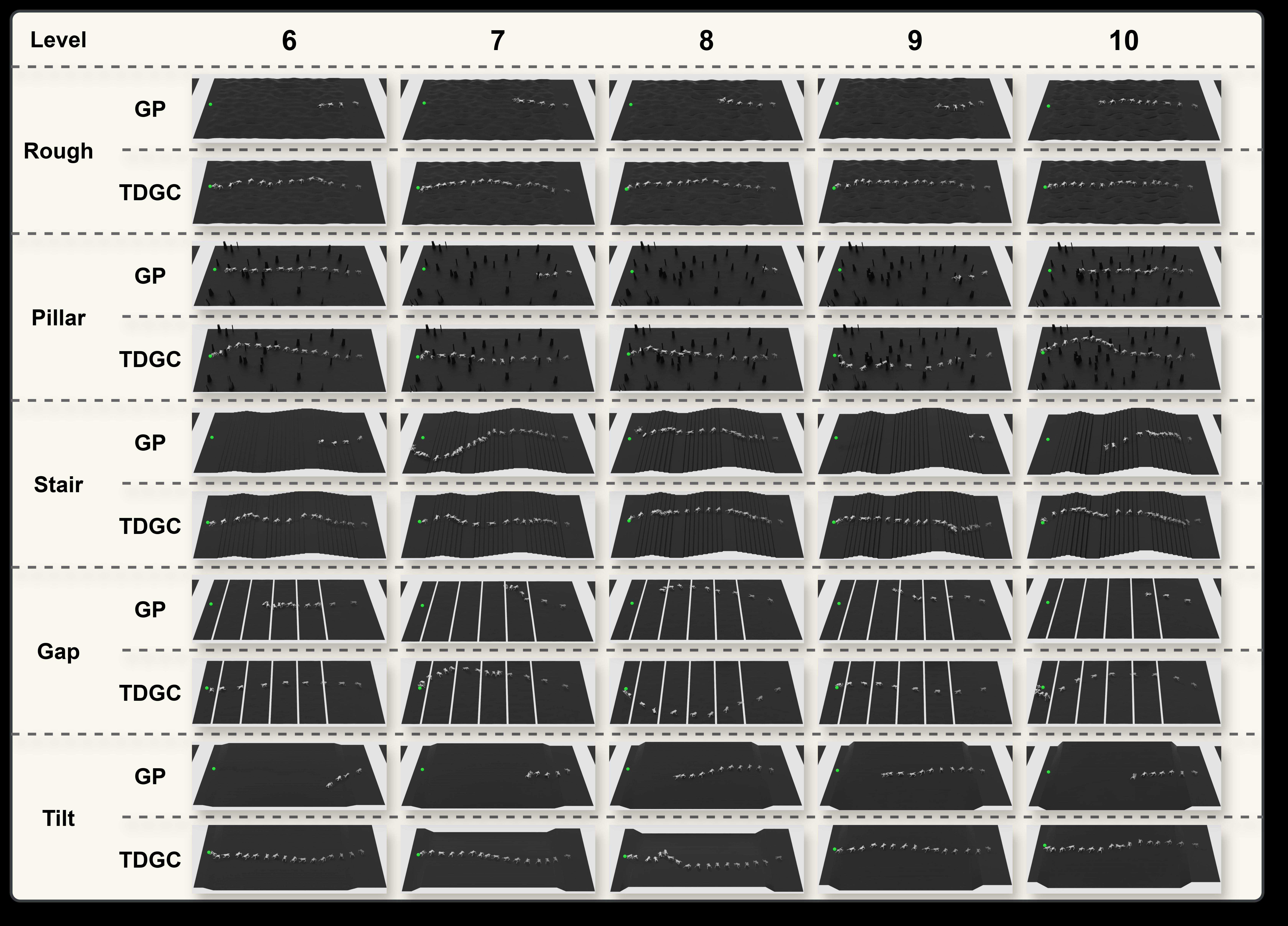}
  \vspace{-19pt}
  \caption{Qualitative rollouts on hard terrain levels (6--10) across five terrain families, comparing GP and TDGC.}
  \label{fig:qual_hard_levels}
  \vspace{-15pt}
\end{figure*}

\begin{table}[t]
  \centering
  \caption{Key training hyperparameters used across all experiments.}
  \label{tab:exp_setup_params}
  \vspace{-0.6em}
  \small
  \setlength{\tabcolsep}{4pt}
  \renewcommand{\arraystretch}{1.15}
  \rowcolors{2}{gray!10}{white}
  \begin{tabular}{p{0.62\columnwidth} p{0.30\columnwidth}}
    \toprule
    \rowcolor{gray!22}
    \textbf{Hyperparameter} & \textbf{Value} \\
    \midrule
    Parallel environments ($N_{\mathrm{env}}$) & 100 \\
    Training budget ($I_{\max}$) & 20000 iterations \\
    Random seed & 42 \\
    Rollout horizon ($T$) & 24 steps/env \\
    Policy optimizer & Recurrent PPO\\
    Learning rate & $1\times10^{-4}$ \\
    Entropy coefficient & $1\times10^{-4}$ \\
    Discount factor ($\gamma$) & 0.99 \\
    GAE parameter ($\lambda$) & 0.95 \\
    PPO clip ($\epsilon$) & 0.2 \\
    Epochs per update & 5 \\
    Mini-batches per epoch & 4 \\
    \bottomrule
  \end{tabular}
  \vspace{-1.0em}
\end{table}

\section{EXPERIMENTS}
This section evaluates the proposed TDGC framework on episodic navigation over procedurally generated mixed terrains. We first describe the experimental setup, then present results.

\subsection{Experimental Setup}
All experiments are conducted in the Isaac Lab physics simulation environment on a GPU accelerated workstation. We use a consistent training configuration across all runs for fair comparison. The hierarchical controller is trained in two stages. We first train the low level gait conditioned executor and then freeze it. The high level task policy is subsequently optimized on top of the frozen executor using a recurrent PPO variant. Parallel simulation is used to improve sample efficiency, and the random seed is fixed for reproducibility. Key training hyperparameters shared across experiments are summarized in Table~\ref{tab:exp_setup_params}.

We evaluate navigation performance on a terrain curriculum grid with five terrain families and multiple difficulty levels per family. To assess robustness under challenging conditions, we focus on the five hardest levels, corresponding to Levels 6 to 10. For each terrain family and each selected level, we run 100 independent evaluation episodes. An episode is considered successful if the robot reaches the goal region at least once during the episode, where a goal reach is detected when the goal distance is smaller than 0.5. The resulting evaluation protocol provides a consistent benchmark for comparing navigation performance across terrain types and difficulty levels. Representative qualitative results are shown in Fig.~\ref{fig:qual_hard_levels}, and the corresponding quantitative results are discussed in the following subsection.

\subsection{Experimental Results}
We evaluate the proposed method, \textbf{TDGC}, on the hardest terrain levels, namely Levels 6 to 10, across all five terrain families. Following the success rate protocol described in Sec.~V-A, we conduct $K=100$ independent episodes for each terrain family at each hard level, resulting in 500 evaluation episodes per terrain family. Across all terrains, \textbf{TDGC} achieves a mean success rate of 87.4\%, which indicates strong and reliable goal reaching performance under challenging terrain geometry and contact conditions.

Fig.~\ref{fig:qual_hard_levels} provides a qualitative comparison between \textbf{TDGC} and the baseline gait policy (\textbf{GP}) on the same set of hard levels. In general, \textbf{TDGC} generates trajectories that are smoother, more coherent, and more consistently directed toward the goal region, whereas GP is more likely to exhibit stalled motions or unstable traversals as terrain difficulty increases. Beyond the overall success rate, we also observe clear and interpretable gait selection patterns induced by the hierarchical decision mechanism. On \textit{Stair} terrains, \textbf{TDGC} often approaches the steps with a lateral body orientation and selects the \textit{trot} gait for ascent. This behavior suggests that a sideways climbing strategy with diagonal leg support can improve stability and foothold placement on terrains with periodic elevation changes. On \textit{Gap} terrains, \textbf{TDGC} frequently traverses cracks by moving backward while selecting the \textit{bound} gait. This behavior indicates a consistent strategy that uses stronger paired leg propulsion to cross support discontinuities and to recover more effectively from partial foothold loss. These results show that \textbf{TDGC} not only improves navigation success on difficult terrains, but also produces an interpretable task to gait decision process that is suitable for inspection, diagnosis, and deployment time adjustment.

\section{CONCLUSIONS}
This paper presented \textbf{TDGC}, a hierarchical policy framework for quadruped navigation that connects task level decision making with gait level locomotion through explicit cross layer interfaces. The high level policy generates task dependent gait tuning commands, while the low level gait conditioned controller ensures robust locomotion. We also introduced a structured curriculum learning strategy that progressively increases terrain difficulty during training to improve efficiency and generalization. Experimental results showed that \textbf{TDGC} achieves higher navigation success rates on mixed and out of distribution terrains. These results demonstrate the value of structured hierarchical control for robust and interpretable quadruped navigation.

\addtolength{\textheight}{-12cm}   








References are important to the reader; therefore, each citation must be complete and correct. If at all possible, references should be commonly available publications.


\bibliographystyle{IEEEtran}
\bibliography{refs}

\end{document}